\documentclass[conference]{IEEEtran}
\usepackage{times}
\usepackage{url}
\usepackage{latexsym}
\usepackage{amssymb}
\usepackage{multirow}
\usepackage{graphicx}
\usepackage{epstopdf}
\usepackage{array}
\usepackage{subfig}
\usepackage{caption}
\usepackage{flushend}
\newcommand{\argmax}{\arg\!\max}
\begin{document}

\title{Semi-supervised Classification for\\Natural Language Processing}

\author{\noindent \textit{Rushdi Shams}\\
Department of Computer Science, University of Western Ontario,
\\
London, ON N6A 5B7, Canada.\\
email: rshams@csd.uwo.ca.}
\maketitle
\begin{abstract}
Semi-supervised classification is an interesting idea where classification models are learned from both labeled and unlabeled data. It has several advantages over supervised classification in natural language processing domain. For instance, supervised classification exploits only labeled data that are expensive, often difficult to get, inadequate in quantity, and require human experts for annotation. On the other hand, unlabeled data are inexpensive and abundant.  Despite the fact that many factors limit the wide-spread use of semi-supervised classification, it has become popular since its level of performance is empirically as good as supervised classification. This study explores the possibilities and achievements as well as complexity and limitations of semi-supervised classification for several natural langue processing tasks like parsing, biomedical information processing, text classification, and summarization.
\end{abstract}


\section{Introduction}
\label{introduction}
Classical supervised methods use labeled data to train their classifier models. These methods are widespread and used in many different domains including natural language processing. The key material used in different natural language processing tasks is text. The number of text, however, is increasing everyday due to the pervasive use of computing. There are more unlabeled than labeled text since data labeling is expensive due to engagement of human for data annotation. The process also consumes time. These difficulties have serious effects on supervised learning since a good fit of a classifier model requires as much labeled data as possible for its training \cite{shih:2003}. 

Semi-supervised learning can be a good means to overcome the aforementioned problems. The basic principle of semi-supervised learning is simple: use both unlabeled and labeled data to generate classifier models. This makes semi-supervised learning substantially useful for a domain like natural language processing because of a good note, unlabeled text is inexpensive, abundant, and more available than labeled text. In addition, empirically, semi-supervised models have good performance records. On many tasks, they are as good as supervised models and in most cases, they are better than cluster-based, unsupervised models \cite{zhu:2009}. However, many of these results are not conclusive and proper care therefore should be taken due to some serious concerns related to semi-supervised learning.

\begin{figure}[b]
  \centering
 \includegraphics[width=0.4\textwidth]{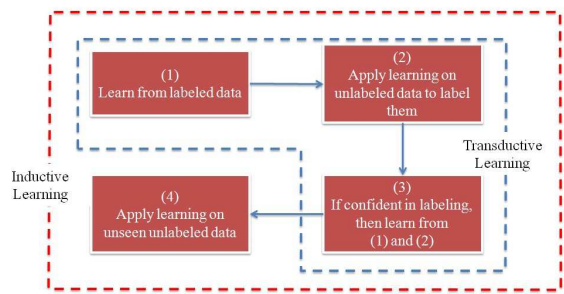}
  \caption{The overview of semi-supervised learning. The figure also outlines the scope of transductive and inductive learning.}
  \label{fig:cycle}
\end{figure}

\begin{figure*}[t]
  \centering
  \subfloat[Supervised decision boundaries for labeled data.]{\label{fig:supervised}\includegraphics[width=.4\textwidth]{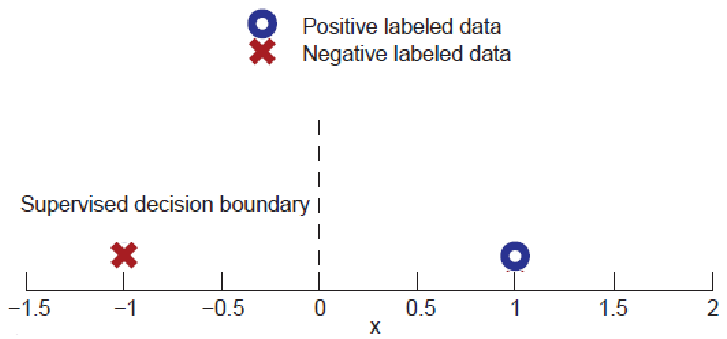}}\quad
  \subfloat[Supervised and semi-supervised decision boundaries for labeled and unlabeled data.]{\label{fig:semisupervised}\includegraphics[width=.4\textwidth]{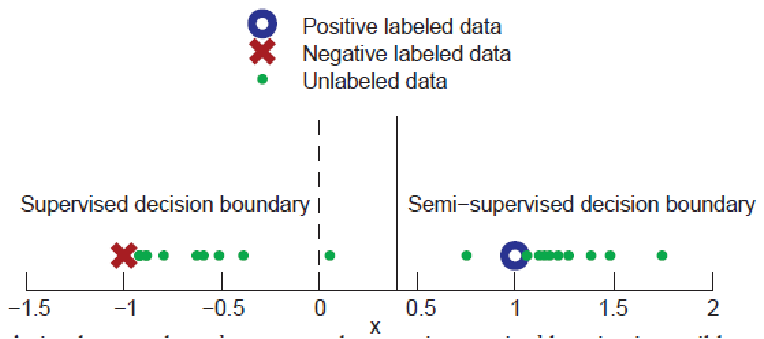}}
  \caption{Supervised and semi-supervised decision boundaries drawn by a random classifier for two labeled and $100$ unlabeled data \cite{zhu:2009}. In \ref{fig:supervised}, the supervised decision boundary is in the middle by averaging the values of the data points. In \ref{fig:semisupervised}, the supervised decision boundary produces more classification errors due to the distribution of the data.}
  \label{fig:decisionboundary}
\end{figure*}

This study explores the use of semi-supervised learning for natural language processing. Interestingly, like traditional machine learning methods, semi-supervised learning can be used to solve classification, regression, and clustering problems. The particular focus of this study, however, is on semi-supervised classification. In this study, popular research papers and classic books are explored to outline the possibilities and achievements of semi-supervised classification for natural language processing. As well, thorough investigations are carried out, both from theoretical and empirical data, to explain the complexity and limitations of this method. Natural language processing is one of the largest research areas of artificial intelligence. The scope of this study is, however, limited to most popular tasks such as parsing, biomedical information processing, text classification, and summarization.

The organization of the paper is as follows. Section \ref{overview} presents an overview of semi-supervised learning that includes learning problems, and different types of semi-supervised algorithms and learning techniques. Following that Section \ref{body} outlines the use of semi-supervised classification for different natural language processing tasks. In Section \ref{conclusions}, several considerations and conclusions are drawn.

\section{Overview of Semi-supervised Learning}
\label{overview}
Unlike supervised and unsupervised learning, semi-supervised learning exploits both labeled and unlabeled data. To start with, semi-supervised methods train models with a very little labeled data. Surprisingly, test results show that marginal labeled data are sufficient to train models with good fit for semi-supervised learning \cite{Chapelle:2010}. The generated models are then applied on unlabeled data in an attempt to label them. The \textit{confidence} of the models in labeling them is measured against a \textit{confidence threshold} set \textit{a priori} by users. Note that learning algorithms often have their own confidence measures that generally depend on their working principles. For instance, class probability values for each data instance are considered as confidence measures for Na\"{i}ve Bayes models \cite{nb:1995}. For an unlabeled data, if the models reach the pre-set confidence threshold, then the newly labeled data are added to the pool of originally labeled data. This process continues unless (i) the models' confidences for the labels stop reaching the threshold, or (ii) the models confidently label all the unlabeled data and there are no unlabeled data remaining in the dataset. The interesting cycle of labeling and re-labeling of semi-supervised learning is illustrated in Figure \ref{fig:cycle}.

\subsection{Learning Problems}
\label{learning problems}
Semi-supervised learning problems can be broadly categorized into two groups: (i) transductive learning and (ii) inductive learning. Transductive learning is like a \textit{take-home exam}. This group of semi-supervised learning tries to evaluate the goodness of a model assumption on unlabeled data after training a classifier with the available labeled data. Inductive learning, on the contrary, is often seen as an \textit{in-class exam}---it evaluates the goodness of a model assumption on \textit{unseen, unlabeled} test data after training a classifier with both labeled and unlabeled data. Figure \ref{fig:cycle} shows the boundaries between these two types of semi-supervised learning. While the entire cycle in the figure illustrates inductive learning, steps $1$---$3$ describe transductive learning.

\subsection{Working Principle}
\label{working principle}
Figure \ref{fig:decisionboundary} can be referred to understand how semi-supervised learning works with a very few labeled but abundant unlabeled data. Figure \ref{fig:supervised} shows that based on the position of a positive ($x=1$) and a negative ($x=-1$) labeled data, a supervised decision boundary is drawn right at $x=0$ based on the average of the data points. However, given only these two labeled data and $100$ unlabeled data (represented by \textit{green dots} in Figure \ref{fig:semisupervised}), this supervised decision boundary still remains at $x=0$. In contrast, had a semi-supervised classifier been used, the boundary would have shifted more to the right (say some point at $x=0.4$) than the supervised decision boundary. This shift is due to the distribution of unlabeled data points considering the position of the positive and negative examples. In this particular case, the semi-supervised classifier assumes that the green dots near to the \textit{red cross} point form one kind of data distribution while the green dots near to the \textit{blue circle} form a different one. Interestingly, semi-supervised learning fails in many intriguing cases, where the distributions of labeled and unlabeled data are not as distinguishable as seen in Figure \ref{fig:decisionboundary}.

\begin{figure}[b]
  \centering
  \subfloat[Joint probability calculated by generative algorithms]{\label{fig:joint}\includegraphics[width=.2\textwidth]{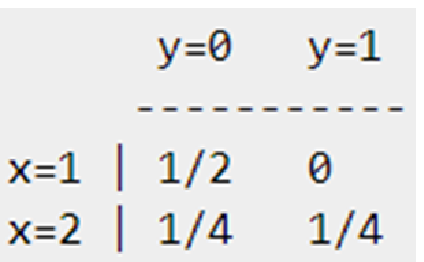}}\quad
  \subfloat[Conditional probability calculated by discriminative algorithms.]{\label{fig:conditional}\includegraphics[width=.2\textwidth]{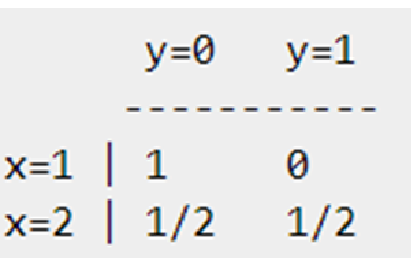}}
  \caption{Probability distribution of the data as seen by generative and discriminative algorithms. Generative algorithms calculate the joint probability distribution of the data while discriminative algorithms deal with their conditional probability.}
  \label{fig:joint-conditional}
\end{figure}

\subsection{Types of Algorithms}
\label{algorithms types}

There are several semi-supervised algorithms and most of them can be categorized into two groups based on their properties: (i) generative algorithms and (ii) discriminative algorithms. The models generated by these two types of algorithms are therefore called generative and discriminative models, respectively. The following can explain the key difference between the two types of models. Say, we are given a set of \textit{speeches} given by human presenters. As well, a set of languages are provided. The task is to simply classify every speech into one of the languages. This learning problem can be solved in either of the following two ways: first, the learner learns each language and then attempts to classify the speeches according to its learning. Second, the learner learns the differences among the speeches according to various attributes or features present in them and then attempts to classify the speeches according to its learning. Note that for the second case, the learner does not need to learn all the languages. The prior is called a \textit{generative} learner and the latter is known as a \textit{discriminative} learner. Let us take a look into these two types of algorithms mathematically. Say, we are given a set of instances $x$ and their classes $y$ in the form of $(x,y)$: $(1,0), (1,0), (2,0), (2,1)$. Generative algorithms attempt to find out the \textit{joint probability}, $p(x,y)$ from these data (see Figure \ref{fig:joint}) while discriminative algorithms calculate their \textit{conditional probability}, $p(x|y)$ (Figure \ref{fig:conditional}). Now, for supervised algorithms a discriminative model predicts the label $y$ from training example $x$ as follows:
\begin{equation}
 f(x)=\argmax_yp(y|x).
 \label{eq:1}
\end{equation}

However, from the Baye's theorem, we know that
\begin{equation}
p(y|x)=\frac{p(x|y)p(y)}{p(x)}.
\label{eq:2}
\end{equation}

However, for Equation \ref{eq:1}, $p(x)$ can be ignored since it finds the function, $f(x)$ for the maximum value of $y$. Therefore, ignoring $p(x)$ in Equation \ref{eq:2} gives us 

\begin{equation}
 f(x)=\argmax_yp(x|y)p(y).
 \label{eq:3}
\end{equation}

Interestingly, Equation \ref{eq:3} is what supervised, generative algorithms use to induce their models. In other words, for supervised algorithms, Equation \ref{eq:1} is used to find out class boundaries based on the given training instances $x$ and Equation \ref{eq:3} is used to generate $x$ for any given $y$. The latter, however, is not found as easily for semi-supervised algorithms as for supervised algorithms. The first and foremost reason for this is that in semi-supervised problems, the algorithms cannot completely ignore $p(x)$ because most of what it has are the distributions of training examples (i.e., $p(x)$). Moreover, for semi-supervised algorithms, a very few class labels are provided (for training examples) and therefore from the few given $y$'s, the conditional probabilities, $p(x|y)$ are difficult to generate. This is a key difference between supervised and semi-supervised algorithms. An example is provided to understand the difference better. For semi-supervised algorithms, Equation \ref{eq:1} can be substituted by 

\begin{equation}
p(y|x)=\frac{p(x|y)p(y)}{\sum_{{y}'}p(x|{y}')p({y}')},
 \label{eq:4}
\end{equation}
where ${y}'$ denotes the classes of the few given training examples $x$. Equation \ref{eq:4} has a probability density function $p(x|y)$ in its numerator. If the distribution of $x$ comes from a  \textit{Gaussian} and it is a function of mean vector and covariance matrix of the \textit{Gaussian}, then using a Maximum Likelihood Estimate, the mean vector and covariance matrix can be tuned to maximize the density function. Thereafter, this \textit{tuning} can be optimized using an \textit{Expectation-Maximization} (\textsc{em}) algorithm. Note that according to the distribution of $x$, different algorithms use different techniques for \textit{tuning} and \textit{optimizing} the density function $p(x|y)$ in Equation \ref{eq:4}. Among the semi-supervised algorithms, Transductive Support Vector Machine (\textsc{tsvm}) and graph-based methods are generative algorithms while \textsc{em} and self-learning are discriminative algorithms.

\subsection{Types of Learning}
\label{learning type}
The semi-supervised learning can be broadly categorized into three: (i) self-training, (ii) co-training, and (iii) active learning. 

\subsubsection{Self-training}
\label{self training}

In self-training, from a set of initially labeled data $L$, a classifier, $C_{1}$ is generated. This classifier is then applied on a set of initially unlabeled data $U$. According to a pre-set \textit{confidence threshold}, the classifications of unlabeled data are observed. If the classifier's confidence reaches the threshold, the newly classified instances are concatenated with $L$ to produce a set $L_{new}$ and removed from $U$ to produce $U_{new}$. A second classifier, $C_{2}$ is generated from $L_{new}$, and thereafter applied on $U_{new}$. This cycle continues until the classifier converges---which means that either (a) all the unlabeled data are confidently labeled by the classifier or (b) the classifier's confidence stops reaching the threshold for several cycles. Self-training is very simple and particularly useful if the supervised algorithm is difficult to modify. Nonetheless, self-training performs poorly for datasets that contain large number of outliers.

\subsubsection{Co-training}
\label{co training}

In contrast to self-training, for co-training, two partitions $L_{1}$ and $L_{2}$ are created from the initially labeled data $L$. The partitions are based on two different sets of attributes or features $V_{1}$ and $V_{2}$ (in semi-supervised literature, they are often referred to as \textit{views}). Then, two classifiers independently generates respective models $F_{1}$ and $F_{2}$ from $L_{1}$ and $L_{2}$ using $V_{1}$ and $V_{2}$. Following that, from the unlabeled data pool $U$, \textit{k} most confident predictions of $F_{1}$ are added to $L_{2}$ and \textit{k} most confident predictions of $F_{2}$ are added to $L_{1}$. These added examples are removed from $U$. $F_{1}$ is re-trained with $L_{1}$ and $F_{2}$ is re-trained with $L_{2}$. This cycle continues until the classifiers converge. Finally, using a voting or averaging method, test data are classified. Note that co-training can be seen as self-training with two or more classifiers. Co-training is very useful if the attributes or features naturally split into two distinguishable sets. However, there are two important conditions that should be met for co-training to work. Given enough labeled data, 
\begin{enumerate}
\item each view alone should be sufficient to make good classifications and
\item the \textit{co-trained} algorithms should individually perform good.
\end{enumerate}

\subsubsection{Active Learning}
\label{active learning}

Finally, for active learning a model is generated from labeled data and attempts to classify unlabeled instances. The classification it makes is then provided to a human expert called \textit{the oracle} for her judgment. The correctly labeled instances according to the oracle are then added to the pool of labeled data while the instances with incorrect labels remain in the unlabeled data pool. This process continues until the unlabeled data pool becomes empty. Active learning is very useful for limited available data (both labeled and unlabeled). Because of the presence of an oracle, this semi-supervised learning is slow and almost always expensive.

\section{Semi-supervised Classification for Natural Language Processing}
\label{body}
In this section, different natural language processing applications of semi-supervised classification are discussed. The discussion is mainly based on the findings from several classic and state-of-the-art literature from the domain of parsing, text classification, text summarization, and biomedical information mining.

\subsection{Parsing}
\label{parsing}

Steedman \textit{et al.} \cite{Steedman:2003} found that self-training has very small effects on parser improvements. Similar results are reported by Clark \textit{et al.} \cite{Clark:2003} who applied self-training to part-of-speech (\textsc{pos}) tagging. The only works that reported successful execution of self-training to improve parsers are very few \cite{reading-parsing:2006} \cite{Bacchiani:2006}. This paper concentrates on the work of McClosky \textit{et al.} because they do not \textit{adapt} the parser in use that because adaptation has some drastic effects on self-training. Rather than using an adaptive parser, the \textit{Charniak} parser used in their research utilized both labeled and unlabeled data that come from the same source domain. Using of a re-ranker besides the parser is also what makes their work different than many contemporary work. The parser uses third order Markov grammar and five probability distributions that are conditioned with more than five linguistic attributes. Firstly, the parser produces $50$-best parses for the sentences in the datasets. Secondly, a \textit{maximum entropy} re-ranker with over a million attributes re-ranks these parses. The experiment is extensive: datasets used in this experiment are Penn treebank section $2-21$ for training (approximately $40,000$ \textit{wall street journal} articles), section $23$ for testing, and section $24$ for held-out data. $24$ million \textit{LA Times} articles were used as unlabelled data collected from the North American News Text Corpus (\textsc{nanc}). The authors experiment with and without the re-ranker as they added unlabelled sentences to their labeled data pool. They found that the parser performs better with the re-ranker system. The improvement reported is about $1.1$\% F-score---among which the self-trained parser contributes $0.8$\% and the re-ranker contributes $0.3$\%). The authors also did some experiments with sentences in section $1$, $22$, and $24$ to see how the self-trained parser performs at sentence level. Each sentence in these sections was labelled as \textit{better}, \textit{no change} or \textit{worse} compared to the baseline F-score for the sentences. Interestingly, the outcomes showed that the parser had improvement neither for unknown words nor for prepositional phrases. However, there was an explicit improvement for intermediate-length sentences but no improvement for the extremes (Goldilocks effect). The parser performs poorly for conjunctions. 

Zhu \cite{Zhu:2006}, however, asserted that in semi-supervised classification, unlabeled sentences for which the parser accuracy is unusually better than normal should be restricted to be included in the pool of labeled data. McClosky \textit{et al.} \cite{reading-parsing:2006}, however, stated that they did not followed this approach particularly. The speed of the semi-supervised Charniak parser is similar to its supervised version but it requires more memory to execute the cycles involved in self-training. Also, the labeled and unlabeled data were collected from two different datasets (although they are both newspaper sources) that usually limits the success of self-training. Nevertheless, the experiment is a success and this question is unanswered in their paper.

\subsection{Text Classification}
\label{text classification}

Semi-supervised classification has been used widely in natural language processing tasks such as spam classification, which is a form of text classification. The results in $2006$ ECML/PKDD spam discovery challenge \cite{Bickel:2006} indicated that spam filters based on semi-supervised classification outperformed supervised filters. Extensive experiments showed that semi-supervised filters work better when source of available labeled examples differs from those to be classified. Interestingly, Mojdeh and Cormack \cite{Mojdeh:2008} found completely different results when they re-designed the challenge with different collections of email datasets.

The $2006$ ECML/PKDD discovery challenge had two interesting tasks. The first task is called the \textit{Delayed Feedback} where the filters are trained with emails $T_{1}$ and then they classify some test emails $t_{1}$. In their second cycle of training, they are trained with $T_{1}$ and $t_{1}$ combined and the training continues for the entire dataset for the challenge. The best ($1-$\textsc{auc}) reported in the challenge is a remarkable $0.01$\%. The second task for the challenge is called the \textit{Cross-user Train} where the classifiers are trained on one particular set of emails and then tested on a completely different set of emails. The best ($1-$\textsc{auc}) reported for this task is greater than the first task: $0.1$\%. The best performing filters in the challenge were all semi-supervised filters and based on support vector machines \textsc{svm} and \textsc{tsvm} \cite{Joachims:1999}, Dynamic Markov Compression (\textsc{dmc}) \cite{bratko:2006}, and Logistic regression with self-training (\textsc{lr}) \cite{Cormack-semi:2006}. On the other hand, in $2007$ TREC Spam Track Challenge \cite{trec-spam:2006}, the participating spam filters were trained with publicly available emails and their model accuracy was tested on emails collected from user inboxes (i.e., personalized emails). In an attempt to see whether semi-supervised filters perform as good as it was reported in \cite{Bickel:2006}, Mojdeh and Cormack \cite{Mojdeh:2008} reproduced the work by replacing the datasets of ECML/PKDD challenge with TREC challenge datasets. The delayed feedback task was reproduced as follows: first, $10,000$ messages were used for training and the next $60,000$ messages were divided into six batches each containing $10,000$ messages. Second, the remaining $5,000$ messages were kept for testing the models. On the other hand, to reproduce the Cross-user Train task, $30,338$ messages from particular user inboxes were used for training while $45,081$ messages from other users were used for model evaluation.

The experimental outcomes showed that for both the tasks, the semi-supervised versions of \textsc{dmc}, \textsc{lr}, and \textsc{tsvm} underperformed for LREC dataset. Their respective $1-$\textsc{auc} scores for the delayed feedback task were $0.090$, $0.046$, and $0.230$. On the other hand, the $1-$\textsc{auc} of their supervised versions were $0.016$, $0.049$, and $0.030$ for the task. For the cross-user task, the $1-$\textsc{auc} of  the semi-supervised \textsc{dmc}, \textsc{lr}, and \textsc{tsvm} filters were $9.97$, $10.72$, and $24.3$, respectively. For the same task, their supervised versions performed way better. The authors also reported a cross-corpus experiment to reproduce the results of ECML/PKDD Challenge. Here, the first $10,000$ messages from the TREC $2005$ dataset were considered. Besides, the TREC $2007$ dataset was split into $10,000$ message segments. The outcomes again showed that self-training is harmful for the filters. Except the \textsc{tsvm} filter, the rest of the two semi-supervised filters failed to perform as good as their supervised versions.

Keeping the aforementioned results in mind, we can say that semi-supervised classification is applicable to text classification but the performance depends on the labeled and unlabeled training data, and the source from which the data are derived.

\subsection{Extractive Text Summarization} 
\label{text summarization}

Wong \textit{et al.} \cite{reading-summary:2008} have conducted a comparative study where they produced extractive summaries by using both supervised and semi-supervised classifiers. The authors used four traditional groups of attributes to train their classifiers: (1) surface (2) relevance (3) event, and (4) content attributes. They tried different combinations of the attributes and found that the classifiers had produced better summaries when the surface, relevance, and content attributes were combined. The novelty of their work is that they used supervised \textsc{svm} as well as its semi-supervised version called probabilistic \textsc{svm} or \textsc{psvm} to generate classifiers and compared their performances. As performance measure they considered \textsc{rouge} scores and found that the \textsc{rouge-i} score of their \textsc{svm} classifier is $0.396$ while the human \textsc{rouge-i} was $0.42$ when compared to the gold standard summaries. On the other hand, the co-training with the \textsc{psvm} and Na\"{i}ve Bayes classifiers produced summaries that have \textsc{rouge-i} of $0.366$. Although this performance is not better than what they found with the supervised \textsc{svm} or human summaries, it was better than supervised \textsc{psvm} and Na\"{i}ve Bayes classifiers. Note that as their datasets, the authors used the \textsc{duc 2001} dataset\footnote{Download at: \url{http://duc.nist.gov}}. The dataset contains $30$ clusters of relevant documents. Each cluster comes with model summaries created by the dataset annotators. $50$, $100$, $200$, and $400$-word summaries are provided for each cluster. Among the clusters, $25$ are used as training data while the remaining $5$ clusters are used for testing. The authors also concluded that the \textsc{rouge-i} scores of their classifier are better if they produce $400$-word summaries for the test clusters. 

Nevertheless, the reported methodology of the paper has some serious drawbacks. Many of the methods used in this research are not in line with what had been found by classic empirical studies. For instance, the co-training is done on the same attribute space that violates the primary hypothesis of co-training: two classifiers used in co-training should use separate views (see Section \ref{co training}). Secondly, the authors selected the set of attributes (surface, relevance, and content attributes) by only considering the performance of \textsc{psvm} with them and ignoring the performance of the supervised Na\"{i}ve Bayes with them.

\subsection{Biomedical Information Mining}
\label{biomedical information mining}

Now-a-days, there is much impetus for information mining from biomedical research papers. Researchers put significant effort to mine secondary information such as protein interaction relations from biomedical research papers to help identify primary information like DNA replication, genotype-phenotype relations, and signaling pathways. The first and foremost task for protein interaction relation miners is to classify sentences in research papers that describe one or more protein interactions. These sentences are called the \textit{candidate sentences}. A number of supervised tools are developed to classify candidate sentences from biomedical articles (see for example \cite{Donaldson:2003}, \cite{Mitsumori:2006}, and \cite{Sugiyama:2003}). However, the first semi-supervised approach for the task was reported by Erkan \textit{et al.} \cite{Erkan:2007}. Their approach identified candidate sentences using similarities of the paths present between two protein names found from the dependency parses of the sentences. What follows are the brief descriptions of their method. The authors produced dependency trees for each sentence from two given datasets. The paths between two protein names in the parse trees were then analyzed. According to these paths, the sentences were labeled and treated as the gold standard for the tool's evaluation. Given the paths, two distance-based measures, cosine similarity and edit distance, were used by their tool to find out interactions between the proteins. These measures were provided to both supervised and semi-supervised algorithms to generate models to classify the sentences in the datasets. The labels predicted by the supervised and semi-supervised classifiers were then evaluated against the gold standard. According to the outcomes, the semi-supervised classifiers performed better than their supervised versions by a wide margin. Four algorithms were used to generate the classifiers among which two are supervised (\textsc{svm} and \textit{K}-Nearest Neighbor (\textsc{knn})) and the rest were their respective semi-supervised versions (\textsc{tsvm} and Harmonic Functions). The distance-based measures were used to generate attributes for the classifiers and were extracted from two datasets named AIMED and Christine-Brun (CB). The AIMED dataset contains $4,026$ sentences of which $951$ describe protein interactions while the CB dataset is composed of $4,056$ sentences of which $2,202$ describe protein interactions. Each of the four algorithms then generated a classifier from the two sets of attributes found from the two distance measures. Experimental outcomes show that for the AIMED dataset, \textsc{tsvm} with edit distance attributes performed the best with $59.96$\% F-score. This F-score was significantly better than the F-scores found using the supervised classifiers. Comparisons showed that the F-score with \textsc{tsvm} was significantly better than those reported by two contemporary work \cite{Mitsumori:2006} \cite{Yakushiji:2005}. On the other hand, the tool performed even better on the CB dataset where its \textsc{tsvm} classifier with edit distance based attributes produced an F-score of $85.20$\%. Similar to the result found with the AIMED dataset, the performances of the supervised classifiers were not satisfactory. The authors also examined the effect of the size of the labeled training data for the classifiers. In the case of AIMED, the authors found that with small labeled training data, semi-supervised algorithms were better. In addition, \textsc{svm} performed poorly with less training data but as more data became available for its training, it started to perform well. On the other hand, for the CB dataset, \textsc{knn} performed poorly with much labeled data. Interestingly, \textsc{svm} performed competitively with the semi-supervised classifiers with more labeled data.

Note that \textsc{tsvm} is susceptible to the distribution of the labeled data. However, the work did not report any test on the data distribution. The AIMED dataset, in addition, has class imbalance problem that seriously affects the performance of \textsc{tsvm} classifiers. This can be seen as the limitation of the work since it did not explain why in their case the \textsc{tsvm} classifier performed better than the rest.

\section{Conclusions}
\label{conclusions}
The findings of empirical research on parsing, text classification, text summarization and biomedical information mining are investigated in this study. According to most of them, semi-supervised classification has substantial advantages over supervised classification when labeled data are difficult to manage and unlabeled data are abundant. This paper also outlines the theories behind the success of semi-supervised classification. According to the theories, there is no \textit{free lunch} for semi-supervised classification rather its success depends on underlying data distribution, data complexity, model assumption, choice of proper algorithm, problem in hand, and most of all---experience. Surprisingly, the investigation has found that the classic studies often do not consider the \textit{do's and don'ts} suggested by the theories. Despite the success reported in the empirical studies, it is therefore inconclusive whether semi-supervised classification can be really as useful as supervised classification.

\begin{figure}[t]
  \centering
 \includegraphics[width=0.3\textwidth]{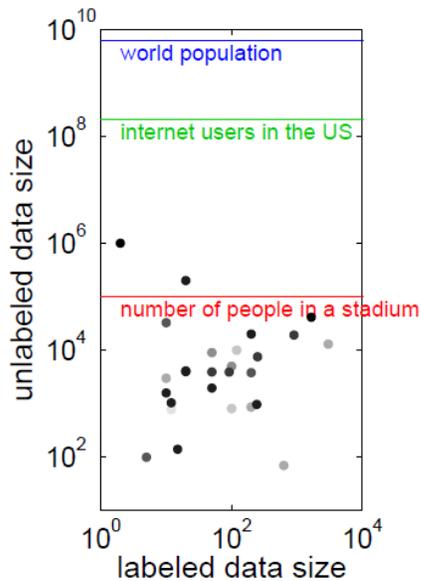}
  \caption{The use of labeled and unlabeled data in semi-supervised classification. A dot represents a paper that uses semi-supervised classification. Light gray dots mean older papers while dark gray dots mean newer papers \cite{Zhu:2006}.}
  \label{fig:use}
\end{figure}

The complexity associated with semi-supervised classification limits its use. This can be seen from the illustration in Figure \ref{fig:use}. It shows the use of labeled and unlabeled data in semi-supervised classification. Each dot in the illustration represents a paper that uses semi-supervised classification. While the light gray dots represent older papers, the dark gray dots represent recent papers. We can come to two conclusions from this data:

\begin{enumerate}
\item there are not much reported work that implement semi-supervised classification and a bulk of the reported work are old and
\item although the main purpose of using semi-supervised classification is the abundance of unlabeled data, the amount of unlabeled data used in research are at most $10^{6}$ so far---in layman's term, which is just above the number of people in a stadium.
\end{enumerate}

Nevertheless, semi-supervised classification is the only option until now to deal the natural language processing problems where there are more unlabeled than labeled data. This study, however, points out the following suggestions for dealing with semi-supervised classification more effectively:

\begin{enumerate}
\item The model assumption for semi-supervised algorithms must match the problem in hand. For instance, if the classes produce well-clustered data then \textit{expectation-maximization} is a good algorithm to choose; if the attribute space can be naturally split into two sets then co-training is preferred; if two points with similar attribute values tend to be in the same class then graph-based method (not discussed in this paper) can be a reasonable choice; if \textsc{svm} performs well on labeled data then \textsc{tsvm} is a natural extension; and given the supervised algorithm is complicated and difficult to modify, self-training is useful.
\item The distributions of both labeled and unlabeled data need to be investigated. \textsc{tsvm}, for instance, performs poorly with unlabeled data that have highly overlapped positive and negative distribution since it assumes that its decision boundary would go right through the densest region. Therefore, in this case a \textsc{tsvm} classifier usually produces a lot of false positives and false negatives.
\item The proportion of labeled and unlabeled data is important to notice before choosing an algorithm. However, there is not conclusive remark on how the proportion affects the overall classification performance.
\item It has been found empirically that there is an effect of dependency among attributes on semi-supervised classification. To be more specific, with fewer labeled examples, the number of dependent attributes should be kept as low as possible.
\item \textit{Data noises} should be investigated as they have effect on classification performance. It is easier to detect noise in the labeled data than in unlabeled data. Note that data noise has less effect on semi-supervised classification than supervised classification.
\item The labeled and unlabeled data, usually, are collected from different sources and this can affect the classification performance. If the labeled and unlabeled data are collected from completely different sources and their properties differ, then rather than using semi-supervised classification, \textit{transfer learning} and \textit{self-taught classification} are encouraged to use \cite{Raina:2007}.
\end{enumerate}

\bibliographystyle{IEEEtran}
\bibliography{rushdibibliography}

\end{document}